\documentclass[a4paper]{llncs}

\usepackage{graphicx}
\usepackage{subcaption}
\usepackage{amssymb}
\usepackage[utf8]{inputenc}
\usepackage{url}
\usepackage{color}
\usepackage{amsmath}
\usepackage{todonotes}
\usepackage{lipsum}  
\usepackage{xargs}          
\usepackage{tablefootnote}
\usepackage{notoccite}
\usepackage{multirow}
\usepackage{lipsum}

\usepackage[acronym]{glossaries}

\makeglossaries 

\newacronym{AutoML}{AutoML}{Automated Machine Learning}
\newacronym{ML}{ML}{Machine Learning}
\newacronym{DSGE}{DSGE}{Dynamic Structured Grammatical Evolution}
\newacronym{SGE}{SGE}{Structured Grammatical Evolution}
\newacronym{GE}{GE}{Grammatical Evolution}
\newacronym{GP}{GP}{Genetic Programming}
\newacronym{BNF}{BNF}{Backus-Naur Form}
\newacronym{CFG}{CFG}{Context-Free Grammar}
\newacronym{RECIPE}{RECIPE}{Resilient Classification Pipeline Evolution}
\newacronym{SVM}{SVM}{Support Vector Machine}
\newacronym{ANN}{ANN}{Artificial Neural Network}
\newacronym{UCI}{UCI}{University of California Irvine}
\newacronym{TPOT}{TPOT}{Tree-based Pipeline Optimization Tool}
\newacronym{EC}{EC}{Evolutionary Computation}

\AtBeginEnvironment{align}{\setcounter{equation}{0}}

\begin{document}

\mainmatter  

\title{Evolution of Scikit-Learn Pipelines with Dynamic Structured Grammatical Evolution}

\author{Filipe Assun\c{c}\~ao\inst{1,2} \and Nuno Louren\c{c}o\inst{1} \and\\ Bernardete Ribeiro\inst{1}\and Penousal Machado\inst{1}}

\authorrunning{Assun\c{c}\~ao et al.}
\titlerunning{Incremental Evolution and Development of Deep Artificial Neural Networks}

\institute{CISUC, Department of Informatics Engineering,\\ University of Coimbra, Coimbra, Portugal\\
\email{\{fga, naml, bribeiro, machado\}@dei.uc.pt} \and LASIGE, Department of Informatics, Faculdade de Ciencias,\\ Universidade de Lisboa, Lisboa, Portugal}

\maketitle

\begin{abstract}
The deployment of \gls{ML} models is a difficult and time-consuming job that comprises a series of sequential and correlated tasks that go from the data pre-processing, and the design and extraction of features, to the choice of the \gls{ML} algorithm and its parameterisation. The task is even more challenging considering that the design of features is in many cases problem specific, and thus requires domain-expertise. To overcome these limitations \gls{AutoML} methods seek to automate, with few or no human-intervention, the design of pipelines, i.e., automate the selection of the sequence of methods that have to be applied to the raw data. These methods have the potential to enable non-expert users to use \gls{ML}, and provide expert users with solutions that they would unlikely consider. In particular, this paper describes AutoML-DSGE -- a novel grammar-based framework that adapts \gls{DSGE} to the evolution of Scikit-Learn classification pipelines. The experimental results include comparing AutoML-DSGE to another grammar-based \gls{AutoML} framework, \gls{RECIPE}, and show that the average performance of the classification pipelines generated by AutoML-DSGE is always superior to the average performance of \gls{RECIPE}; the differences are statistically significant in 3 out of the 10 used datasets.

\end{abstract}

\glsresetall

\section{Introduction}
\label{sec:intro}

\begin{figure}[t!]
    \centering
    \includegraphics[width=\linewidth]{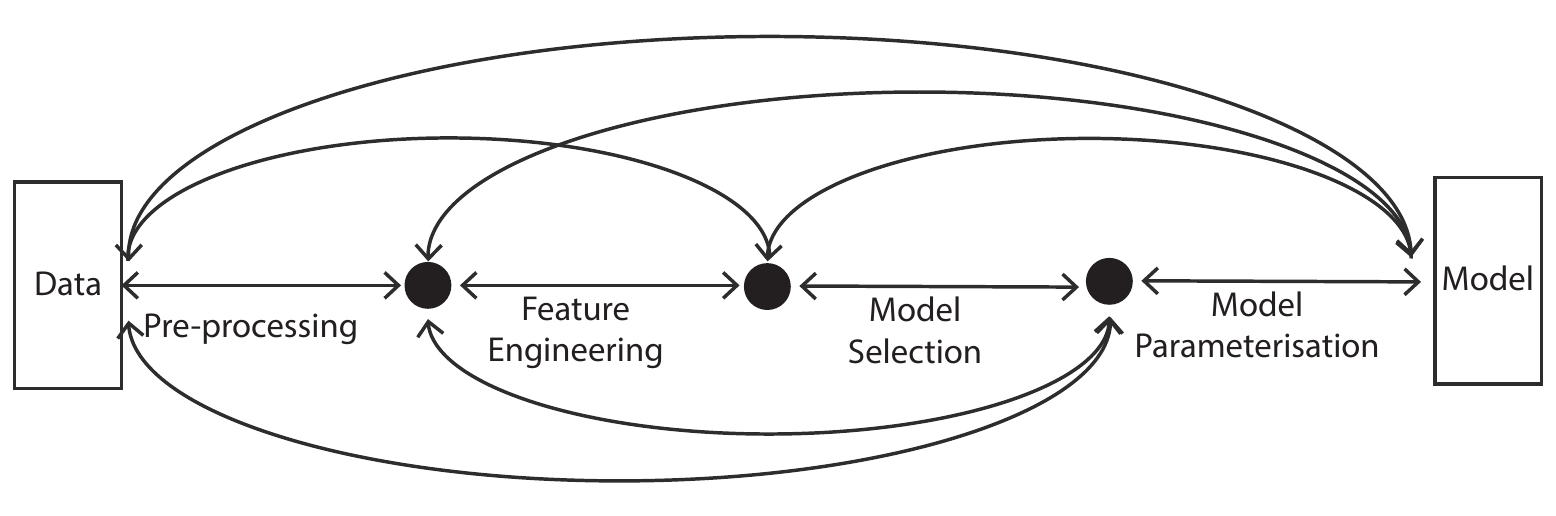}
    \caption{From data to ML model deployment.}
    \label{fig:ml_human}
\end{figure}

Nowadays, with the ever-growing amount of collected information the challenge is not concerned with the lack of information, but rather on how to design efficient \gls{ML} models that can extract useful knowledge, or aid in the automation of daily-life tasks. Typically, to deploy a \gls{ML} system we need to follow a pre-defined number of steps: (i) pre-process the data; (ii) design, extract, and select features, i.e., data characteristics; (iii) select the most appropriate \gls{ML} model; and (iv) parameterise the \gls{ML} model. The flow of the iterative steps that one must traverse from the data to the model is depicted in Figure~\ref{fig:ml_human}: the multiple steps are all interconnected, which means that for example in case we have already a model, but we acknowledge that the set of features is not the most adequate one we may be thrown back to the beginning of the process again. In addition, even when the practitioner is a \gls{ML} expert, and is well aware of which models are more adequate for particular tasks, it still needs to design features, which are domain-dependent, and therefore often require domain-expertise, and sometimes multidisciplinary teams.   

To overcome the difficulty caused by the correlation between the multiple choices that have to be made prior to deploying a \gls{ML} system we can resort to \gls{AutoML}. In brief words \gls{AutoML} concerns searching for the most effective \gls{ML} models for a particular task. One of the key-advantages of \gls{AutoML} is that it does not require human input, and consequently the gain is twofold: (i) on the one hand it empowers non-expert users with the ability to apply \gls{ML} models to their problems; (ii) on the other hand, it opens the door to novel solutions, that a human-expert would potentially neglect. 

The current work focuses on \gls{AutoML} applied to classification datasets. The common approach of \gls{AutoML} to this sort of problems is to evolve a classification pipeline, i.e., an ordered sequence of tasks that are performed to accurately distinguish between the different classes of the problem. The pipeline tasks can be any known form of data pre-processing; feature design, extraction, or selection; or \gls{ML} algorithm. In particular, we evolve Scikit-Learn~\cite{scikit-learn} pipelines with \gls{DSGE}~\cite{dsge}. Our main contributions are:
\begin{itemize}
    \item The proposal of a new grammar-based \gls{AutoML} framework based on \gls{DSGE}: AutoML-DSGE;
    \item The release of the framework as open-source, available on GitHub: \url{https://github.com/fillassuncao/automl-dsge};
    \item The performance of a wide set of experiments on multiple classification tasks;
    \item The comparison of AutoML-DSGE to previous \gls{AutoML} methods. The results show that the results of AutoML-DSGE are always superior to those reported by other grammar-based AutoML methods, and are statistically superior in 3 out of the 10 used datasets. 
\end{itemize}

The remainder of the paper is structured as follows. Section~\ref{sec:background_sota} surveys multiple \gls{AutoML} methods; Section~\ref{sec:dsge} describes \gls{DSGE}; Section~\ref{sec:automl_dsge} details the evolution of Scikit-Learn classification pipelines with \gls{DSGE}; Section~\ref{sec:experimentation} analyses the experimental results; and Section~\ref{sec:conclusions} draws conclusions and addresses future work. 

\section{Related Work}
\label{sec:background_sota}

The most common and widely used form of \gls{AutoML} is grid search: the best parameterisation of a \gls{ML} model is discovered by an exhaustive search of all the combinations of a grid of parameters. However, grid search suffers from the curse of dimensionality, i.e., the explosion in the number of parameters drastically increases the amount of setups that need to be tested. To deal with the previous we can instead use grid search methods that seek to narrow the number of setups, for example by adapting the resolution of the grid in run-time~\cite{DBLP:journals/ijon/JimenezLD09a}. Nonetheless, grid search has the advantage that it is highly parallelisable. To overcome the issue of having to explore the entire grid of hyper-parameters we may instead apply random search. While grid search performs an exhaustive enumeration of the domain, random search selects the combinations of the hyper-parameters in a stochastic manner.  Random search is as parallelisable as grid search;.Nonetheless, it is non-adaptive~\cite{DBLP:conf/sc/YoungRKLP15}, and with very high dimensional search spaces it also struggles to find near-optimal solutions. According to Bergstra et al., given the same computational time, random search is able to discover better parameterisations for \glspl{ANN} than grid search~\cite{DBLP:conf/nips/BergstraBBK11,DBLP:journals/jmlr/BergstraB12}.

An alternative to grid and random search are Bayesian methods~\cite{ShahriariSWAF16}, which model probabilistically the behaviour of the system, in order to drive search towards regions of the domain that are prone to generate good parameterisations. Snoek et al. applied Bayesian optimisation to tune the parameters of the Branin-Hoo function, Logistic Regression, Online Linear Discriminant Analysis, Latent Structured Support Vector Machines, and Convolutional Neural Networks~\cite{DBLP:conf/nips/SnoekLA12}. Bergstra et al. have demonstrated that statistical methods can perform better at hyper-parameter optimisation~\cite{DBLP:conf/icml/BergstraYC13} than manual tuning or random search. Other class of heuristic approach is \gls{EC}, which has also been widely used to optimise \gls{ML} algorithms (e.g., \cite{chunhong2004automatic,DBLP:journals/ijon/FriedrichsI05}).

The majority of the methods mentioned until now focus on the optimisation of a specific \gls{ML} model. Nonetheless, the ultimate goal of \gls{AutoML} is to fully automate the entire process: from the data pre-processing, and feature design and selection up to the model choice and parameterisation. Recently, there have been competitions seeking to promote such systems; an example is  ChaLearn~\cite{DBLP:conf/icml/GuyonCEEJLMRRSS16}. The challenge is organised into 6 increasingly difficult levels (preparation, novice, intermediate, advanced, expert, and master), where the ultimate goal is to ``create the perfect black box eliminating the human in the loop''~\cite{DBLP:conf/ijcnn/GuyonBCEEHMRSSV15}.

Weka~\cite{DBLP:books/sp/datamining2005/FrankHHKP05} and Scikit-learn~\cite{scikit-learn} are examples of two \gls{ML} libraries that enable users to explore their data and easily deploy learning models. They make available stable implementations of the vast majority of \gls{ML} methods, but despite providing default parameterisation they are not suit for effectively solving all problems. Auto-WEKA~\cite{DBLP:conf/kdd/ThorntonHHL13,DBLP:journals/jmlr/KotthoffTHHL17}, \gls{TPOT}~\cite{DBLP:conf/gecco/OlsonBUM16}, Hyperopt-Sklearn~\cite{komer2014hyperopt}, Auto-Sklearn~\cite{DBLP:conf/nips/FeurerKESBH15}, and \gls{RECIPE}~\cite{de2017recipe}, are examples of methods that aim at evolving the pipelines for the Weka and Scikit-learn libraries, from the pre-processing of the raw data to the parameterisation of the model to be used (in essence they automate the flow-chart depicted in Figure~\ref{fig:ml_human}). Except for \gls{TPOT} and \gls{RECIPE}, all the previous methodologies are based on Bayesian optimisation; \gls{TPOT} and \gls{RECIPE} use \gls{GP}. The goal is to search for Weka or Scikit-Learn pipelines, i.e., sequences of the libraries' primitives that perform feature selection and classification. The frameworks are not only responsible for selecting the primitives but also promote their parameterisation. Auto-Weka, Hyperopt-Sklearn, Auto-Sklearn and \gls{RECIPE} generate pipelines of fixed size; \gls{TPOT} allows the generation of pipelines of unrestricted size, i.e., it does not have a fixed number of pre-processors, and multiple copies of the dataset can be used in simultaneous, so that multiple methods are applied to it, and then the features combined. Whilst the majority of these approaches target the maximisation of the classification performance, in addition \gls{TPOT} also seeks for compact pipelines.

The focus of the current work is on \gls{AutoML} approaches based on \gls{EC}. In particular, we are interested in grammar-based methods, such as \gls{RECIPE}. The main advantage of grammar-based methods over others is that they facilitate the definition of the search space, and thus in case we have a-priori knowledge about the problem we can bias the grammar. On the other hand, the grammar enables the framework to be easily extended: to add more methods to the search space we just require the definition of new production rules. To the best of our knowledge, \gls{RECIPE} is the only grammar-based \gls{AutoML} framework that aims at optimising classification pipelines. The current paper introduces AutoML-DSGE and compares it to \gls{RECIPE}. AutoML-DSGE is based on \gls{DSGE}, which is detailed next.

\section{Dynamic Structured Grammatical Evolution}
\label{sec:dsge}

To properly introduce \gls{DSGE}~\cite{dsge} we must start by detailing \gls{SGE}~\cite{lourencco2016unveiling}, which is a variant of \gls{GE}~\cite{o2001grammatical}. The three methods are grammar-based \gls{GP} approaches, and thus the search space is defined by means of a \gls{CFG}. \glspl{CFG} are rewriting systems, and thus the grammar, G, can be formally defined by a 4-tuple $G=(N,T,P,S)$, where: (i) $N$ is the set of non-terminal symbols; (ii) $T$ is the set of terminal symbols; (iii) $P$ is the set of production rules of the form $x ::= y$, $x \in N$ and $y \in \{N \cup T\}^*$; and (iv) $S$ is the start symbol (or axiom). An example of a \gls{CFG} is shown in Figure~\ref{fig:cfg_example}. The main difference between the methods lies on the encoding of the individuals, and thereby on the genotype decoding procedure. 

The individuals in \gls{GE} are encoded as linear ordered sequences of integers; each integer represents a derivation step and is called a codon. The genotype to phenotype mapping works by reading the codons sequentially, from left to right. Starting from the axiom the mapping procedure iteratively decides which production rule should be applied to expand the leftmost non-terminal symbol. To select the production rule the modulo mathematical operation (\%) is used to find the remainder after the division of the codon by the number of possibilities for expanding the leftmost non-terminal symbol. The remainder defines the expansion possibility that should be applied to the leftmost non-terminal symbol. No codon is read when there is only a possibility for expanding a non-terminal symbol. On the other hand, grammars can be recursive, and thus the number of codons may be insufficient; in such cases the sequence of codons is re-used from the start (wrapping). To avoid entering an infinite wrapping loop, or generating solutions that are too complex to be evaluated, a maximum number of wrappings is set, and when this bound is reached the mapping procedure is halted, and the individual is assigned the worst possible fitness value. 

The drawbacks commonly pointed to \gls{GE} are low locality and high redundancy~\cite{DBLP:conf/eurogp/KeijzerORC02,DBLP:conf/ppsn/ThorhauerR14}. The locality measures how the changes in the genotype impact the phenotype. In \gls{GE} there is not a one-to-one mapping between the codons and the non-terminal symbols, and therefore it is easy for a change in one of the codons to affect all the derivation steps from that point on-wards (low locality). On the other hand, the redundancy is concerned to the fact that in \gls{GE} it is possible that different genotypes generate the same phenotype because of the modulo operation used on the decoding procedure. 

\gls{SGE} solves the limitations of \gls{GE} by introducing a new genotypic representation that defines a one-to-one mapping between the codons and the non-terminal symbols, i.e., instead of a single ordered sequence of codons the genotype is composed by multiple independent ordered sequences of codons, one for each non-terminal symbol. The size of each sequence of codons is of the maximum number of possible expansions for the non-terminal symbol it encodes, and thus there is no wrapping. The use of the modulo operation is not required as we know exactly which non-terminal symbol the codon encodes. 

In \gls{SGE} the genotypes encode more codons than the ones used in the decoding procedure, and consequently the genetic operators may easily act upon non-coding genes. This under some circumstances can slow down evolution. To prevent this effect, the genotype of \gls{DSGE} is similar to those of \gls{SGE} with one main difference: it only encodes the codons strictly required for decoding the individual. In case mutations affect the amount of necessary codons, the genotype is expanded. In this paper we use \gls{DSGE}; the code for \gls{DSGE} can be found in the GitHub repository \url{https://github.com/nunolourenco/sge3}.

\section{AutoML-DSGE}
\label{sec:automl_dsge}

The goal of this paper is to introduce a new framework, to which we call AutoML-DSGE, that adapts \gls{DSGE} to the evolution of classification pipelines. In particular, we optimise Scikit-Learn~\cite{scikit-learn} pipelines. Next, we define pipelines (Section~\ref{sec:pipeline}), the used grammar (Section~\ref{sec:grammar}), and detail the evolution of pipelines using \gls{DSGE} (Section~\ref{sec:representation}). The code for AutoML-DSGE is released as open-source software, and can be found in the GitHub repository \url{https://github.com/fillassuncao/automl-dsge}.

\begin{table}[t!]
    \centering
    \begin{tabular}{c|c|c}
      \bf Pre-processing & \bf Feature manipulation      & \bf Classification   \\ \hline
        Imputer          & VarianceThreshold             & ExtraTreeClassifier          \\
        Normalizer       & SelectPercentile              & DecisionTreeClassifier       \\
        MinMaxScaler     & SelectFpr                     & GaussianNB                   \\
        MaxAbsScaler     & SelectFwe                     & BernouliNB                   \\
        RobustScaler     & SelectFdr                     & MultinominalNB               \\
        StandardScaler   & RFE                           & SVC                          \\
                         & REFCV                         & NuSVC                        \\
                         & SelectFromModel               & KNeighborsClassifier         \\
                         & IncrementalPCA                & RadiusNeighborsClassifier    \\
                         & PCA                           & NearestCentroid.             \\
                         & FastICA                       & LDA                          \\
                         & GaussianRandomProjection      & QDA                          \\
                         & SparseRandomProjection        & LogisticRegression           \\
                         & RBFSampler                    & LogisticRegressionCV         \\
                         & Nystroem                      & PassiveAggressiveClassifier  \\
                         & FeatureAgglomeration          & Perceptron                   \\
                         & PolynomialFeatures            & Ridge                        \\
                         &                               & RidgeCV                      \\
                         &                               & AdaBoostClassifier           \\
                         &                               & GradientBoostingClassifier   \\
                         &                               & RandomForestClassifier       \\
                         &                               & ExtraTreesClassifier         \\
    \end{tabular}
    \vspace{7pt}
    \caption{Scikit-Learn classes that are allowed to be part of the pipelines.}
    \label{tab:pipeline}
\end{table}

\subsection{Pipelines}
\label{sec:pipeline}

In the field of \gls{ML} a classification pipeline is defined as an ordered set of operations that are performed to the data instances in order to accurately separate them in the multiple classes of the dataset. The operations in the pipeline can be grouped into 3 disjoint sets: (i) data pre-processing; (ii) feature design and selection; and (iii) classification. Table~\ref{tab:pipeline} enumerates the methods that are considered to form the pipelines in the current work. Recall that we focus on classification pipelines, and thus only classification algorithms are taken into account. Nonetheless, the extension of the approach to regression algorithms is straight-forward. We will optimise Scikit-Learn pipelines, and thus the methods in the table are Scikit-Learn implementations. Further details can be found in \url{https://scikit-learn.org/stable/user_guide.html}.

\subsection{Grammar}
\label{sec:grammar}

\begin{figure}[t!]
    \centering
    \footnotesize
    \begin{align}
        {<}\text{pipeline}{>} ::= & \, {<}\text{preprocessing}{>} \, {<}\text{algorithm}{>}  \\
                   & \, | \, {<}\text{algorithm}{>} \\
        {<}\text{preprocessing}{>} ::= & \, {<}\text{imputation}{>} \, | \, {<}\text{bounding}{>} \, | \, {<}\text{dimensionality}{>} \, | \, \\
                   & \, | \, {<}\text{binarizer}{>} \, | \, {<}\text{imputation}{>} \, {<}\text{bounding}{>} \\
                   &  \, | \, {<}\text{imputation}{>} \, {<}\text{binarizer}{>} \\
                   & \, | \, \ldots \\
        {<}\text{imputation}{>} ::= & \, \text{preprocessing:imputer} \, {<}\text{strategy\_imp}{>} \\
        {<}\text{strategy\_imp}{>} ::= & \, \text{strategy:mean} \, | \,  \text{strategy:median} \, | \, \text{strategy:most\_frequent} \\
        \ldots & \\
        \ldots & \\
        {<}\text{algorithm}{>} ::= & \, {<}\text{strong}{>} \, | \,  {<}\text{weak}{>} \, | \, {<}\text{tree\_ensemble}{>} \\
        \ldots & \\
        \ldots & \\
        {<}\text{weak}{>} ::= & \, {<}\text{nearest}{>} \, | \,  {<}\text{discriminant}{>} \, | \, \ldots \\
        \ldots & \\
        \ldots & \\
        {<}\text{nearest}{>} ::= & \, \text{classifier:radius\_neighbors} \, {<}\text{radius}{>} \, {<}\text{weights}{>} \\
                             & \, {<}\text{k\_algorithm}{>} \, {<}\text{leaf\_size}{>} \, {<}\text{p}{>} \, {<}\text{d\_metric}{>} \\
        {<}\text{radius}{>} ::= & \, \text{radius:RANDFLOAT(1.0,30.0)} \\
        {<}\text{weights}{>} ::= & \, \text{weights:uniform} \, | \, \text{weights:distance} \\
        {<}\text{k\_algorithm}{>} ::= & \, \text{algorithm:auto} \, | \, \text{algorithm:brute} | \ldots \\
        {<}\text{leaf\_size}{>} ::= & \, \text{leaf\_size:RANDINT(5,100)} \\
        \ldots & \\
        \ldots & 
    \end{align}
    \caption{CFG used by AutoML-DSGE for optimising Scikit-Learn pipelines.}
    \label{fig:cfg_example}
\end{figure}

The grammar used by AutoML-DSGE describes the search space of the Scikit-Learn classification pipelines. The grammar is shown in Figure~\ref{fig:cfg_example}. The production rules are only partially shown because of space constraints: the grammar is comprised of 89 production rules that encode the different pipeline methods and their parameterisation. The complete grammars can be found in \url{https://github.com/fillassuncao/automl-dsge/tree/master/sge/grammars}. There is a separate grammar for each dataset because of specific dataset parameters, e.g., number of features. The used grammars are adapted from the grammars used by \gls{RECIPE}, which is the method we compare AutoML-DSGE to. 

The axiom of the grammar is the pipeline non-terminal symbol, and consequently the pipeline can be found by either pre-processing and classification methods (line 1) or just by the classification method (line 2). The current version of AutoML-DSGE does not consider ensembles. The extension of AutoML-DSGE to enable the optimisation of ensembles could be easily introduced by adding a recursive production rule to build pipelines with more than one classifier algorithm, each a voter of the ensemble. The pre-processing methods manipulate the dataset and features (lines 3-6), and the classification methods cover a wide range of \gls{ML} approaches, amongst which, are clustering methods, \glspl{SVM}, trees, or \glspl{ANN} (lines 11-18). In more detail, the pipeline methods are encoded as follows: the pre-processing and classification methods are encoded respectively by the preprocessing and classifier tags, that are placed before the method name (e.g., classifier:radius\_neighbors in line 17). The method name must match the name of the function that is used in the mapping from the phenotype to the Scikit-Learn interpretable code (see Section~\ref{sec:representation}). The same rationale is applied to the method parameters, where the parameter name precedes the parameter value. The parameters can be of three types: (i) closed choice, e.g., the weights parameter, in line 20, that can assume the values uniform or distance; (ii) random integer, e.g., the leaf size parameter in line 22; or (iii) random float, e.g., the radius parameter in line 19.

The search space of AutoML-DSGE, i.e., the number of possible combinations of the grammar is greater than $9.39 \times 10^{17}$. The continuous parameters can generate an infinite number of possibilities, and thus are not considered in the search space size. In addition, the parameters related to the number of features are also not taken into account because they are problem dependent.

\subsection{Evolution of Pipelines}
\label{sec:representation}


The pipelines are evolved using \gls{DSGE}, and therefore, a population of individuals is continuously evolved throughout a given number of generations, until a stop criteria is met. Each individual encodes a different pipeline. The core of the representation of the individuals in AutoML-DSGE is similar to the representation scheme used in \gls{DSGE}, with one main difference related to the need to directly keep real values in the genotype. Otherwise, they would have to be encoded by production rules, such as: 
\begin{align*}
        {<}\text{randfloat}{>} ::= & \, {<}\text{signal}{>} \, {<}\text{rec-number}{>}\, . {<}\text{rec-number}{>}  \\
        {<}\text{signal}{>} ::= & \, - \, | \, + \\
        {<}\text{rec-number}{>} ::= & \, {<}\text{number}{>} \, | \, {<}\text{number}{>} \, {<}\text{number-recursive}{>} \\
        {<}\text{number}{>} ::= & \, \text{0} \, | \, \text{1} \, | \, 2 \, | \, 3 \, | \, 4 \\
                                & \, 5 \, | \, 6 \, | \,7 \, | \, 8 \, | \, 9
\end{align*}
The encoding of real values by means of production rules has two main disadvantages. On the one hand it enlarges the search space. On the other hand there is no easy way to control the limits (minimum and maximum) of the generated real values. In case the search space encompasses two or more real values with different ranges there would be the need for different production rules, one for each real value range. Because of the aforementioned we encode the integers and floats directly, as real values. When expanding the grammar, when we reach a terminal symbol that is either RANDINT or RANDFLOAT we store a tuple in the genotype. The tuple has the format (rand-type, rand-min, rand-max, rand-value), where rand-type can assume integer of float,  the rand-min and rand-max are the minimum and maximum limits of the range, and the rand-value is the randomly generated value of the type rand-type, and within the [rand-min, rand-max] range. The tuple is necessary for performing the mutation, i.e., when a mutation is applied to an individual and it is required to generate a new random value for a specific parameter we must know its type and allowed range. 

\begin{table}[t!]
    \centering
    \begin{tabular}{c|c|c|c|c|c}
        \textbf{Dataset}        & \textbf{\#Inst.}       &  \textbf{\#Feat.} & \textbf{Feat. types} & \textbf{\#Classes} & \textbf{Missing}  \\ \hline
        Breast Cancer           & 699                   & 9                 &  Integer             & 2                  & Yes                    \\ 
        Car Evaluation          & 1728                  & 5                 &  Categorical         & 4                  & No                     \\ 
        Caenorhabditis Elegans  & 478                   & 765               &  Binary              & 2                  & No                     \\
        Chen-2002               & 179                   & 85                &  Real                & 2                  & No                     \\
        Chowdary-2006           & 104                   & 182               &  Real                & 2                  & No                     \\
        Credit-G                & 1000                  & 20                &  Real / Categorical  & 2                  & No                     \\
        Drosophila Melanogaster & 119                   & 182               &  Real                & 2                  & No                     \\
        DNA-No-PPI-T11          & 135                   & 104               &  Real / Categorical  & 2                  & Yes                    \\
        Glass                   & 214                   & 9                 &  Real                & 7                  & No                     \\ 
        Wine Quality-Red        & 1599                  & 11                &  Real                & 10                 & No                     \\ 
        
    \end{tabular}
    \vspace{7pt}
    \caption{Description of the used datasets.}
    \label{tab:datasets}
\end{table}

\gls{DSGE} is a grammar-based approach, and thus the genotype is completely separate from the phenotype. The phenotype does not directly represent a trainable pipeline. Consequently, for assessing the fitness of the individuals we have to perform two sequential steps: (i) map the genotype to the phenotype; (ii) map the phenotype to Scikit-Learn interpretable model. 
To map the genotype to the phenotype the decoding procedure of \gls{DSGE} is adapted: the only difference lies in the decoding of the real-values, where the value in the last position of the tuple is read. The phenotype of AutoML-DSGE is readable, despite not being Scikit-Learn executable code. The readability of the phenotype is facilitated by the fact that each parameter has the parameter name associated to the value; an example of a phenotype is ``classifier:random\_forest criterion:gini max\_depth:None n\_estimators:50 min\_weight\_fraction\_leaf:0.01 \ldots''.

To map the phenotype to a Scikit-Learn interpretable pipeline we have to traverse the phenotype linearly from left to right and for each pre-processing or classifier method create the corresponding Scikit-Learn object. Therefore, for each method in the grammar we have to build a function that creates the Scikit-Learn object: the function receives all the parameters that are encoded in the grammar and outputs the Scikit-Learn object. Whenever the grammar is extended to include more methods we have to create the corresponding functions.

To evaluate the evolved pipelines we use cross-validation (with 3 folds). In the current paper the fitness is the average of the performances on the cross-validation. The metric used to evaluate the performance is the F-measure. We decide for this metric because some of the datasets where we will be conducting the experiments are highly unbalanced. 

The goal of AutoML is to generate (automatically) effective Scikit-Learn classification pipelines that non-expert \gls{ML} users can deploy in their problems and domains. With this in mind, similarly to other approaches, we limit the train time of each pipeline to a maximum CPU time, that in this paper is set to five minutes. For the same reason evolution is halted when there is no improvement for five generations.

\section{Experimentation}
\label{sec:experimentation}

To investigate the ability of AutoML-DSGE to generate effective classification Scikit-Learn pipelines we apply it to the classification of 10 datasets, which are described in Section~\ref{sec:datasets}. The experimental setup is detailed in Section~\ref{sec:exp_setup}, and the analysis of the evolutionary results, and comparison to the pipelines generated by \gls{RECIPE} is carried out in Section~\ref{sec:exp_results}.

\subsection{Datasets}
\label{sec:datasets}

To enable a fair comparison between AutoML-DSGE and \gls{RECIPE} we conduct the experiments on the same datasets used by \gls{RECIPE}: 10 datasets -- 5 from the \gls{UCI} \gls{ML} repository~\cite{uci}, and 5 from bio-informatics~\cite{chen2002gene,chowdary2006prognostic,wan2015predicting}. A summary of the dataset characteristics is shown in Table~\ref{tab:datasets}. The table provides information on the number of instances (\#Inst.), number of features (\#Feat.), type of features (Feat. types), number of classes (\#Classes), and if there are or not missing values in the dataset (Missing).

\begin{table}[t!]
    \centering
    \begin{tabular}{c | c}
        \textbf{Parameter} & \textbf{Value} \\ \hline
        Number of runs & 30 \\ 
        Number of generations & 100\\
        Population size & 100 \\
        Mutation rate & 10\% \\
        Crossover rate & 90\% \\
        Elitism & 5 individuals \\
        Tournament size & 2\\
        Max. pipeline train time & 5 minutes \\
        Max. \#generations without improvement & 5 \\
    \end{tabular}
    \vspace{7pt}
    \caption{Experimental parameters.}
    \label{tab:exp_parameters}
\end{table}

\subsection{Experimental Setup}
\label{sec:exp_setup}

The parameters required for performing the experiments contained in this article are described in Table~\ref{tab:exp_parameters}. The parameters are the same for AutoML-DSGE and \gls{RECIPE}. The maximum CPU train time is measure in minutes, and thus it is important to mention that the experiments are performed in a dedicated server with an Intel(R) Core(TM) i7-5930K CPU @ 3.50GHz, and 32 GB of RAM. 

The code used for AutoML-DSGE, and \gls{RECIPE} can be found, respectively, in the GitHub repositories \url{github.com/laic-ufmg/Recipe/}, and \url{github.com/fillassuncao/automl-dsge}. The code of \gls{RECIPE} was modified to include the evolution stop criteria based on a maximum number of generations without improvement, which despite described in the framework paper~\cite{de2017recipe}, is not included in the current code version. 

To enable the comparison of results we apply the same dataset partitioning scheme used in \gls{RECIPE}: all datasets are split using a 10-fold cross-validation strategy; and thus as we perform 30 evolutionary runs each fold is kept as the test set three times, and the remaining used for training the pipelines. During each run, the test set is kept aside from evolution, and the train set is used to train the pipelines with cross-validation (3 folds). By the end of evolution, the best pipeline is trained using all the train data and applied to the test set. The evolution is conduced using the grammar of Figure~\ref{fig:cfg_example}.

To establish the pair-wise comparison of the results, and check whether or not the differences between AutoML-DSGE and \gls{RECIPE} are statistically significant we use the Wilcoxon Signed-Rank test, with a significance level of 5\%. Further, for the statistically significant differences we compute the effect size. 

\subsection{Experimental Results}
\label{sec:exp_results}

\begin{table}[t!]
    \centering
    \begin{tabular}{c | c | c | c}
        \bf Dataset                 & \bf AutoML-DSGE           & \bf RECIPE                & \bf p-value           \\ \hline
        Breast Cancer               & \bf 0.9568 $\pm$ 0.0296   & 0.9311 $\pm$ 0.0798       & \textbf{0.0264} (++)  \\
        Car Evaluation              & \bf 0.9964 $\pm$ 0.0068   & 0.9962 $\pm$ 0.0079       & 0.9761              \\
        Caenorhabditis Elegans      & \bf 0.6140 $\pm$ 0.0644   & 0.6049 $\pm$ 0.0681       & 0.7948                \\
        Chen-2002                   & \bf 0.9451 $\pm$ 0.0413   & 0.9292 $\pm$ 0.0618       & 0.3371                \\
        Chowdary-2006               & \bf 0.9970 $\pm$ 0.0163   & 0.9812 $\pm$ 0.0514       & 0.0679                \\
        Credit-G                    & \bf 0.7400 $\pm$ 0.0370   & 0.7075 $\pm$ 0.0359       & \textbf{0.0008} (+++) \\
        Drosophila Melanogaster     & \bf 0.6679 $\pm$ 0.1001   & 0.6353 $\pm$ 0.1518       & 0.2585                \\
        DNA-No-PPI-T11              & \bf 0.7114 $\pm$ 0.1194   & 0.7021 $\pm$ 0.0761       & 0.9681                  \\
        Glass                       & \bf 0.7628 $\pm$ 0.1095   & 0.7325 $\pm$ 0.1021       & 0.0524                \\
        Wine Quality-Red            & \bf 0.6600 $\pm$ 0.0387   & 0.6430 $\pm$ 0.0422       & \textbf{0.0257} (++)  \\
    \end{tabular}
    \vspace{7pt}
    \caption{AutoML-DSGE, and RECIPE comparative performance. The results are averages of 30 independent runs.}
    \label{tab:results}
\end{table}

To compare the pipelines generated by AutoML-DSGE and \gls{RECIPE} we conduct evolution for the same datasets, and using equivalent grammatical formulations, i.e., the search space is the same for both frameworks. The test performance (f-measure), for each dataset is presented in Table~\ref{tab:results}. The results are averages of 30 independent runs. A f-measure marked in bold indicates the approach that reports the highest average performance. In addition, the table also reports the p-values for the pair-wise comparisons between the two approaches, and bold p-values indicate statistically significant differences. The effect-size is denoted in brackets after the p-value, with +, ++, and +++ denoting small (0.1 $\leq$ r $<$ 0.3), medium (0.3 $\leq$ r $<$ 0.5), and large (r $\geq$ 0.5) effect sizes, respectively. 

The analysis of the results indicates that AutoML-DSGE reports results that are always superior to those obtained by \gls{RECIPE}. In addition to the higher average, the standard deviation is lower in the AutoML-DSGE results in 7 out of 10 datasets, i.e., for the considered datasets AutoML-DSGE generates more consistently higher results. These differences are statistically significant in 3 datasets (Breast Cancer, Credit-G, and Wine Quality-Red). The effect size is medium twice, and high once. AutoML-DSGE is never worse than \gls{RECIPE}.

The results of Table~\ref{tab:results} report the average performance of the 30 evolutionary runs, for each dataset. Nonetheless, as we are optimising \gls{ML} methods we investigate the generalisation ability of the generated pipelines. To this end, we compute the average difference between the evolutionary, and test performances for the 10 datasets. Except for the Chowdary-2006 and Car datasets, the average difference between the evolutionary and test performance is lower in AutoML-DSGE than in \gls{RECIPE}. Considering all datasets, the average difference between the evolutionary and test set performance is of approximately 0.0328 in AutoML-DSGE and of 0.0589 in \gls{RECIPE}. This proves that the tendency to overfit is lower in AutoML-DSGE, as it reports more often than \gls{RECIPE} evolutionary performances that are closer to the test ones.


\begin{figure}[t!]
    \centering
    \includegraphics[width=\linewidth]{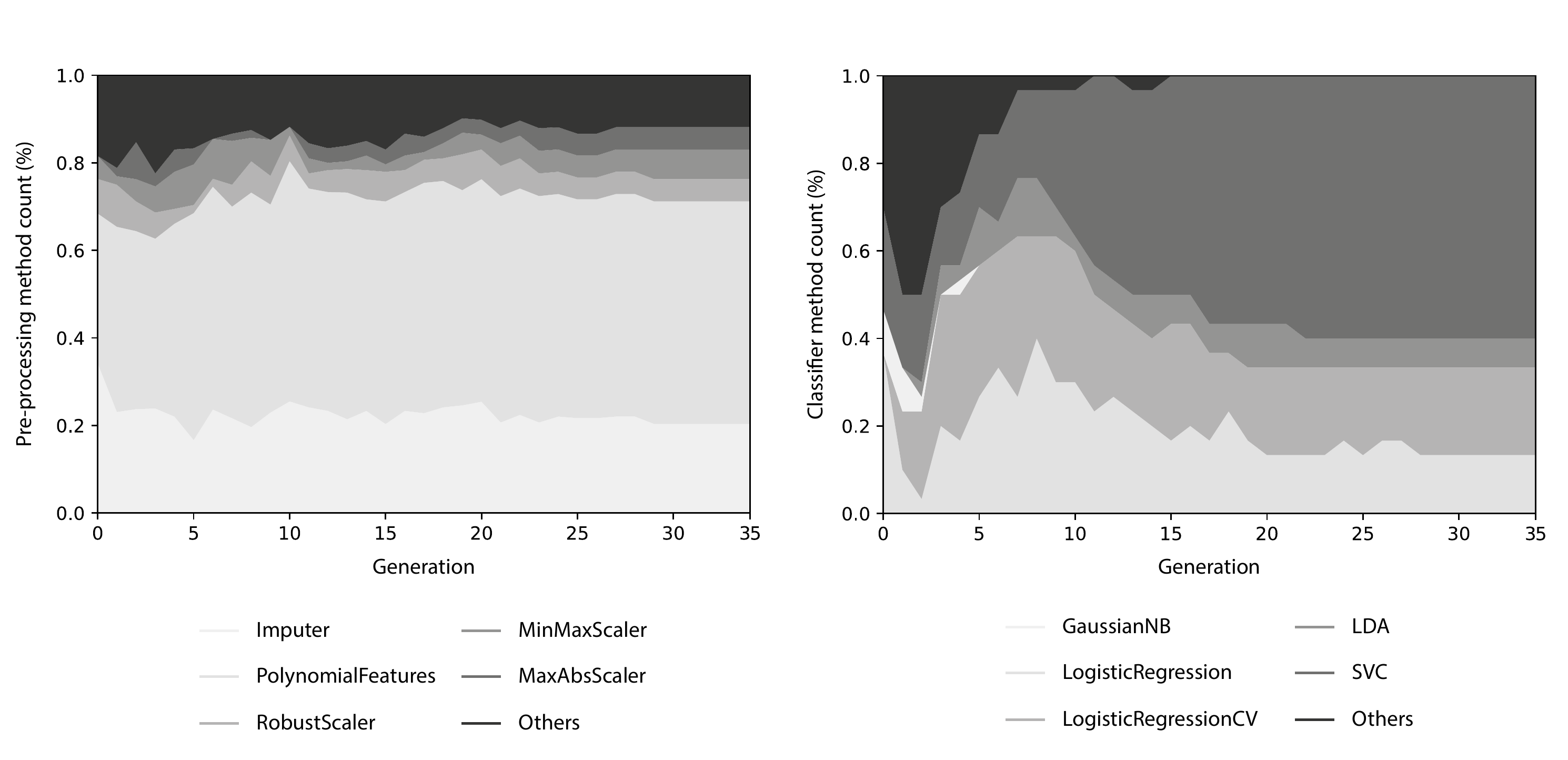}
    \caption{Stacked area charts of the AutoML-DSGE evolution of the pre-processing (left) and classification (right) methods on the Car dataset. The results reflect the percentage of the best pipelines that use each of the methods.}
    \label{fig:pipeline_evolution}
\end{figure}

\begin{figure}[t!]
    \centering
    \includegraphics[width=0.45\linewidth]{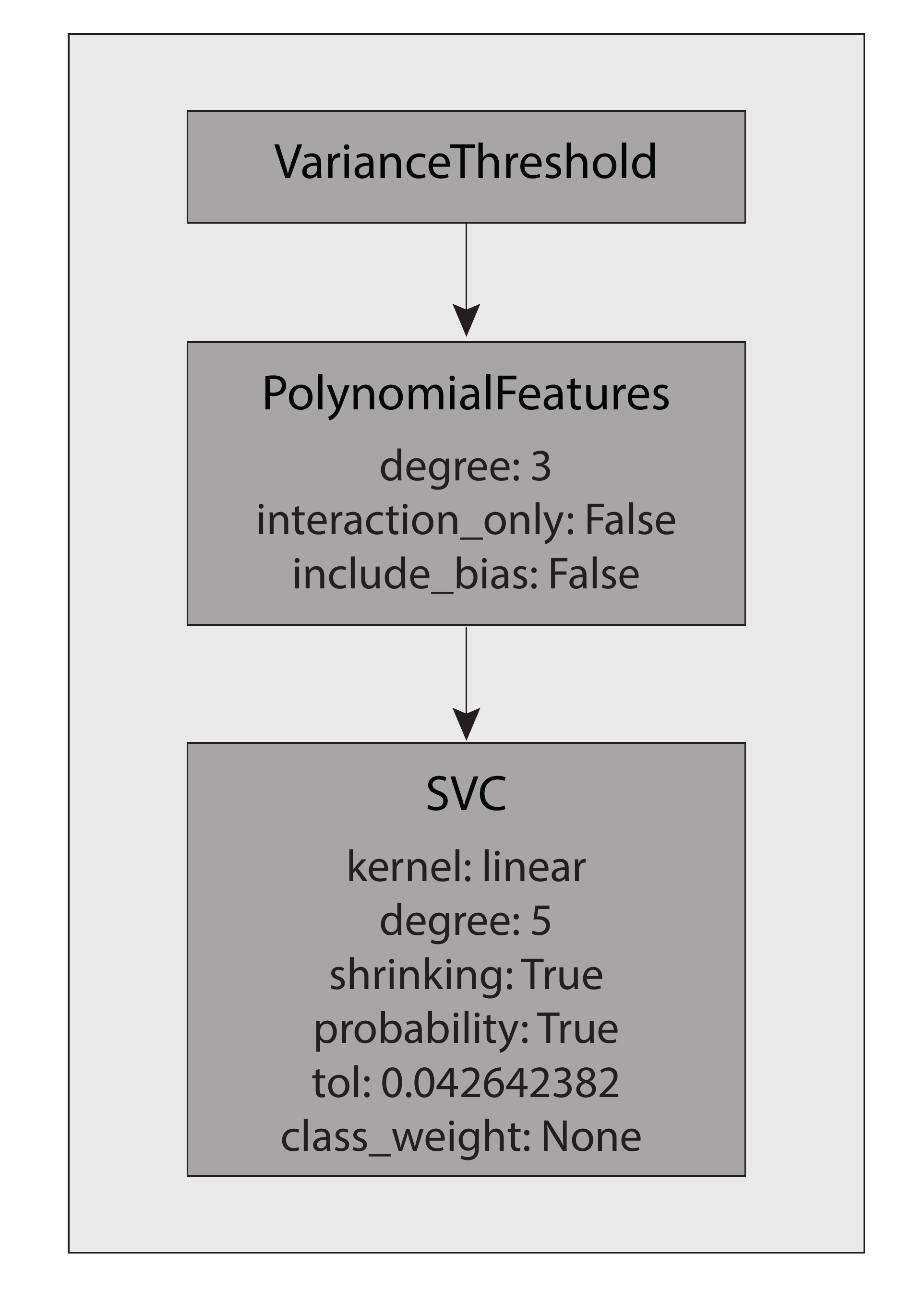}
    \caption{Best pipeline generated for classifying the Car dataset. Each box represents a pipeline method and its parameterisation.}
    \label{fig:best_pipeline}
\end{figure}

To analyse the structure of the pipelines evolved by AutoML-DSGE we inspect the methods that compose them. Due to space constraints we focus on the Car dataset, as it is the dataset where, on average, more generations are performed. Figure~\ref{fig:pipeline_evolution} shows the evolution of the pre-processing and classification methods of the best individuals as generations proceed. The results show the evolution of the percentage of the runs that use each of the pre-processing and classification methods. Recall that the different evolutionary runs can differ in the number of performed generations, and therefore to avoid a misleading representation of the evolution of the methods that compose the pipelines we consider that all runs have the same number of generations. That is,  we consider that all runs evolve for the same number of generations as the longer run (in this case 35 generations). For the evolutionary runs that perform less generations we keep the last generation (which is the best found solution) for the remainder of the generations. The results show that, for the Car dataset, the pre-processing methods distribution does not change as evolution proceeds. On the other hand, a different behaviour is noticeable on the classifier methods, that converge to the SVC, and LogisticRegression (or LogisticRegressionCV) method. The evolution also shows that evolution is focused on the methods that are more effective for that specific dataset. Otherwise, the used methods would be more diverse, and the percentage of the Others would be higher. In particular, we plot in Figure~\ref{fig:best_pipeline} the best pipeline found for classifying the Car dataset. We also inspect the evolutionary patterns in the remaining datasets and acknowledge similar conclusions. It is however important to point out that for the different datasets evolution focuses on different pre-processing and classification methods.

Ultimately, AutoML-DSGE generates no invalid pipelines. After investigating the pipelines that were assigned with the worse possible fitness we conclude that their train is halted because they are unable to train in the the maximum granted CPU time of five minutes, or because they run out of memory.

\section{Conclusions and Future Work}
\label{sec:conclusions}

Prior to the deployment of a \gls{ML} model there are a number of choices that have to be made. There is the need to pre-process the dataset, design, extract and select features, and decide which \gls{ML} model is the most adequate. On top of that, all this sequential choices are correlated, meaning that one affects multiple others. The choices that have to be made require both domain-specific, and \gls{ML} expertise. In an effort to facilitate the widespread use of \gls{ML} models we introduce a novel \gls{AutoML} framework: AutoML-DSGE. 

AutoML-DSGE is a grammar-based \gls{AutoML} approach, and thus the search space is defined in a human-readable \gls{CFG}. This key-point of the framework enables the easy adaptation of AutoML-DSGE to tackle different problems using a wide set of methods. Further, it eases the introduction of a-priori knowledge in the search and tuning of the pipelines. The current version of the framework focuses on the optimisation of Scikit-Learn classification pipelines. The code is released as open-source software, and can be found in the GitHub repository: \url{https://github.com/fillassuncao/automl-dsge}.

We compare the performance AutoML-DSGE to \gls{RECIPE}, which to the best of our knowledge is the only grammar-based AutoML framework. The methods are compared on 10 datasets from different domains. The results show that the pipelines generated by AutoML-DSGE surpass in performance the ones obtained by \gls{RECIPE}; the average performances of AutoML-DSGE are always superior to \gls{RECIPE}, and are statistically superior in 3 datasets (with medium and large effect sizes). Moreover, AutoML-DSGE is less prone to overfitting than \gls{RECIPE}.

Future work will be divided into 4 independent research lines: (i) apply AutoML-DSGE to a wider set of benchmarks; (ii) extend the framework to regression problems; (iii) introduce ensembling and stacking methods; and (iv) enable the user to select between the Weka and Scikit-Learn \gls{ML} libraries, or even own implemented methods.

\section*{Acknowledgments}

\noindent This work is partially funded by: Funda\c{c}\~ao para a Ci\^encia e Tecnologia (FCT), Portugal, under the PhD grant SFRH/BD/114865/2016, and the project grant DSAIPA/DS/0022/2018 (GADgET). We also thank the NVIDIA Corporation for the hardware granted to this research.

\bibliography{bibliography}
\bibliographystyle{splncs03}

\end{document}